\documentclass[letter,10pt,conference]{ieeeconf}
\usepackage{graphicx,amsmath,amssymb,amsfonts,bm,cite,xcolor,algorithmic,textcomp}

\IEEEoverridecommandlockouts
\newcommand{\slm}{$\Sigma \Lambda$~}

\newcommand{\refeqn}[1]{{\eqref{#1}}} %

\newcommand{\refig}[1]{{Fig.~\ref{#1}}} %

\newcommand{\stkout}[1]{\ifmmode\text{\sout{\ensuremath{#1}}}\else\sout{#1}\fi}

\definecolor{MyGreen}{rgb}{0,0.5.2}
\definecolor{MyBlue}{rgb}{0,0.2,1.0}
\definecolor{MyRed}{rgb}{1,0.2,0.2}
\definecolor{MyPurple}{rgb}{0.5,0,1.0}
\definecolor{MyOrange}{rgb}{1.0,0.82,0}
\definecolor{MyBrown}{rgb}{0.65,0.35,0}
\definecolor{MyMagenta}{rgb}{1.,0.0,1.}
\definecolor{MyGrey}{rgb}{0.5,0.5,0.5}
\definecolor{MyLightGrey}{rgb}{0.9,0.9,0.9}

\newcommand{\dani}[1]{{\color{MyGreen}{(Daniel: #1)}}}
\newcommand{\daniel}[1]{{\color{MyGreen}{(Daniel: #1)}}}
\newcommand{\rejean}[1]{{\color{MyRed}{(Rejean: #1)}}}
\newcommand{\ffl}[1]{{\color{MyBrown}{(FFL: #1)}}}
\newcommand{\sylvain}[1]{{\color{MyOrange}{(Sylvain: #1)}}}

\newcommand{\revise}[1]{{\color{MyBrown}{(Revise: #1)}}}

\newcommand{\previous}[1]{{\color{MyGrey}{(Previously: #1)}}}
\newcommand{\reviewer}[1]{{\color{MyBrown}{(Reviewer: #1)}}}

\newcommand{\addition}[1]{{\color{MyBlue}#1}}
\newcommand{\fixed}[1]{{\color{MyGreen}[Fixed: #1]}}
\newcommand{\deleted}[1]{{\color{MyRed}\stkout{#1}}}
\newcommand{\todo}[1]{{\color{MyRed}(\textbf{TODO:} #1)}}

\ifthenelse{\isundefined{\comments}}
{
\renewcommand{\dani}[1]{}
\renewcommand{\daniel}[1]{}
\renewcommand{\rejean}[1]{}
\renewcommand{\sylvain}[1]{}
\renewcommand{\ffl}[1]{}
\renewcommand{\revise}[1]{}
\renewcommand{\previous}[1]{}
\renewcommand{\reviewer}[1]{}

\renewcommand{\addition}[1]{#1}
\renewcommand{\fixed}[1]{}
\renewcommand{\deleted}[1]{}
\renewcommand{\todo}[1]{}

}
{}

\title{\LARGE \bf
Image-driven Robot Drawing with Rapid Lognormal Movements
}
\author{Daniel Berio$^{1}$, Guillaume Clivaz$^{2}$, Michael Stroh$^{3}$, Oliver Deussen$^{3}$\\ R\'ejean Plamondon$^{4}$, Sylvain Calinon$^{2}$ and Frederic Fol Leymarie$^{1}$
\thanks{$^{1}$Goldsmiths, University of London, United Kingdom.
        {\tt\small daniel.berio@gold.ac.uk, ffl@gold.ac.uk}}%
\thanks{$^{2}$Idiap Research Institute and École Polytechnique Fédérale de Lausanne (EPFL), Switzerland.
        {\tt\small guillaume.clivaz@idiap.ch, sylvain.calinon@idiap.ch}}%
\thanks{$^{3}$University of Konstanz, Germany.
        {\tt\small od@uni.kn, michael.stroh@uni-konstanz.de}}%
\thanks{$^{4}$Ecole Polytechnique de Montr\'eal, Canada.
        {\tt\small rejean.plamondon@polymtl.ca}}%
}

\begin{document}

\maketitle
\thispagestyle{empty}
\pagestyle{empty}

\begin{abstract}
  Large image generation and vision models, combined with differentiable rendering technologies, have become powerful tools for generating paths that can be drawn or painted by a robot. However, these tools often overlook the intrinsic physicality of the human drawing/writing act, which is usually executed with skillful hand/arm gestures. Taking this into account is important for the visual aesthetics of the results and for the development of closer and more intuitive artist-robot collaboration scenarios. We present a method that bridges this gap by enabling gradient-based optimization of natural human-like motions guided by cost functions defined in image space.
To this end, we use the sigma-lognormal model of human hand/arm movements, with an adaptation that enables its use in conjunction with a differentiable vector graphics (DiffVG) renderer. We demonstrate how this pipeline can be used to generate feasible trajectories for a robot by combining image-driven objectives with a minimum-time smoothing criterion. We demonstrate applications with generation and robotic reproduction of synthetic graffiti as well as image abstraction.

\end{abstract}

\section{Introduction}

  Collaborative robots are increasingly becoming a consumer technology, making them accessible to both professional and amateur artists as tools for creating physical artworks such as paintings, drawings, and calligraphy \cite{schaldenbrandFRIDACollaborativeRobot2022,Tresset2013}. Unlike standard machines like printers and plotters, articulated robotic arms introduce both new possibilities and challenges, such as the ability to mimic skilled, dynamic brushstrokes and adapt to various tools and surfaces.

In parallel, while AI-driven image processing systems are becoming a widely adopted tools by both artists and designers, physical and embodied AI (a.k.a. robots) currently remains a largely unexplored territory. Recent advances in differentiable vector graphics (DiffVG) are enabling to bridge this gap by extending deep learning image-generation systems from pixel-only outputs to vector-based representations. Specifically, this enables gradient-based optimization of curve and stroke parameters using image-based objectives \cite{schaldenbrandStyleCLIPDrawCouplingContent2022,FransClipdraw2022,xingDiffSketcherTextGuided2024,vinkerCLIPassoSemanticallyAwareObject2022}, ranging from pixel-wise similarity to a reference image to complex objectives driven by the features of large vision and image generation models.

While the outputs of current DiffVG systems can be executed by a robot, these methods typically rely on parametric curve representations \cite{farinCurvesSurfacesCAGD2001} that do not account for the kinematics of the gestures a human would use to produce such strokes. At the same time, drawing, painting, and writing are human motor activities that often require rapid and skillful arm movements. These movements are learned through extensive practice and their kinematic qualities influence how ink or material is deposited on a canvas or a surface, ultimately affecting the aesthetic perception of the resulting trace \cite{Chamberlain2021}.

In this paper, we show how DiffVG technology can be used to directly optimize the parameters of a computational model of human handwriting and drawing movements. We show how this can be used to constrain the generation of drawing gestures with a minimum-time smoothness criterion \cite{Tanaka2006} that is consistent with motor control principles, while also enabling the generation of motions that are kinematically feasible by a robot. This is especially useful for robotic mark-making applications where rapid and human-like movement is desirable.

To describe the kinematics of hand and arm movements, we rely on the sigma-lognormal model \cite{Berio2020Lognormality}, which describes hand/arm movements as the 
superposition of aiming sub-movements, each characterized with a speed profile mathematically modeled using a lognormal function. We provide a differentiable formulation of the model that can be easily integrated into existing DiffVG stroke generation pipelines. The output of our system produces trajectories that do not require an intermediate re-parameterization step and are feasible for a robot by only scaling and resampling according to given velocity and acceleration limits. We demonstrate how this is useful with three practical applications that combine a minimum-time trajectory smoothing criterion with a generation task: (i) fitting trajectories to images of graffiti tags, (ii) generating ``pseudo'' graffiti, and (iii) stylizing images by using a pre-trained multimodal model.

Our focus on reproduction of graffiti stems from the premise that it represents a form of ``urban calligraphy'' \cite{Anssi2015}, where various stylizations and abstractions are applied to letterforms. Graffiti developed on the ``moving canvas'' of the New York City subway in the 1960s \cite{kimvall2014gword} and its aesthetics depend on the execution of rapid and skillful movements, making it a particularly interesting art form to investigate in a robotic setting.

\section{Background and related work}
Differentiable rendering \cite{katoDifferentiableRenderingSurvey2020} enables the propagation of gradients through the transformation of 2D and 3D parametric primitives into pixels. In the 2D domain, DiffVG rendering \cite{liDifferentiableVectorGraphics2020} supports most primitives in the Scalable Vector Graphics (SVG) format, a capability that has facilitated the development of a wide variety of vector graphics generation methods that leverage the power of large pretrained image generation models such as stable diffusion \cite{jainVectorFusionTexttoSVGAbstracting2022} or multimodal text-vision models such as Contrastive Language–Image Pretraining (CLIP)
\cite{radfordLearningTransferableVisual2021}. This has enabled a variety of applications ranging from image vectorization \cite{maLayerwiseImageVectorization2022}, to stylization and abstraction \cite{iluzWordAsImageSemanticTypography2023,vinkerCLIPassoSemanticallyAwareObject2022}, as well as sketch generation \cite{xingDiffSketcherTextGuided2024,FransClipdraw2022,ganzCLIPAGGeneratorFreeTexttoImage2024}.

The DiffVG method currently supports piecewise quadratic and cubic  B\'ezier curves, a representation that, without additional constraints, does not provide high-order continuity that characterizes human movements \cite{flashCoordinationArmMovements1985}. Indeed, most existing sketch/drawing generation in this domain optimize the control points of many short curve segments. Schaldenbrand et al. \cite{schaldenbrandFRIDACollaborativeRobot2022} developed a custom neural rasterizer that takes into account the sim-to-real gap when executing brushstrokes with a robot. However, their method only supports combinations of short brush strokes. In this paper, we demonstrate how an alternate representation can be used to generate long continuous paths that are similar to the ones that would be typically done by a human when drawing or writing.  

Human hand and arm movements are intrinsically smooth and appear to follow a series of ``invariants'' \cite{rosenbaum2009human}. Speed and curvature of human movements tend to show an inverse relation, which for certain types of movements takes the form of a power law \cite{Viviani1991,Plamondon1998}. Intuitively, high curvature requires higher precision and leads to slower movement \cite{Todorov_Minimal_2002}. \addition{Movement time appears to be approximately independent of movement extent, a principle that is commonly referred to as \emph{isochrony} \cite{Jordan1999,Thomassen1985Time}. } Human movements are smooth \cite{balasubramanianAnalysisMovementSmoothness2015} and appear to follow optimality principles \cite{Engelbrecht2001}. This has been explicitly modeled with cost functions such as minimizing the square magnitude of high-order positional derivatives \cite{flashCoordinationArmMovements1985}, 
as well as movement duration \cite{Tanaka2006}. Target-directed movements are characterized by a bell-shaped speed profile, and more complex movements can be described as a space-time superposition of multiple such sub-movements \cite{morasso1981spatial}. The \textit{Kinematic Theory of Rapid Human Movements} (Kinematic Theory for short) has been developed by Plamondon et al. over the years \cite{Plamondon1995}, with various applications in handwriting analysis and generation \cite{Plamondon2014}, as well as computer-aided design of calligraphy and graffiti \cite{BerioEG2018}.

Similarly to our work, Wolniakowski et al. \cite{Wolniakowski2021} reproduce \slm kinematics with an industrial robot, while Berio et al. \cite{Berio16IROS} use the properties of the \slm model to incrementally adjust and reproduce graffiti-like trajectories with a 7 DoF compliant robot. We adopt a similar extension of the \slm model, enabling 2.5D motions with respect to the drawing pad. We use the vertical direction to model brush width and to vary tool pressure. A similar approach to brush width modeling is used by Fujioka et al. \cite{Fujioka2006}, who use uniform B-splines with a minimum jerk cost to reproduce Japanese calligraphy with a robot. A number of works have been developed for robotic reproduction of portraits with a variety of tools \cite{Calinon2005,Low2022,Scalera2019,Song2018}, a goal that we share in this paper. Over a series of works, Tresset et al. \cite{Tresset2013} use a low-cost planar robot arm for pen-based robotic portraits that are made with rapid pen-strokes that are incrementally added in a visual-feedback loop. Deussen et al. \cite{Deussen2012}, Stroh et al. \cite{Stroh2023}, and Schaldenbrand et al. \cite{schaldenbrandFRIDACollaborativeRobot2022} also use a visual feedback loop to place painterly brush strokes for robot painting. While we do not cover visual feedback scenarios in this work, our method is potentially applicable in this setting.
While most of these works produce drawings consisting of many strokes, Singhal et al. \cite{singhal2020chitrakar} generate one-line portraits with a robot by creating paths with ``TSP-art'' \cite{kaplan2005tsp}, a method that combines weighted Voronoi sampling \cite{secordWeightedVoronoiStippling2002} of an image and paths generated by solving a traveling salesman problem (TSP). We use a similar strategy as an initialization for our optimization procedure.

\begin{figure*}[t]
\centering
\includegraphics[width=0.99\textwidth]{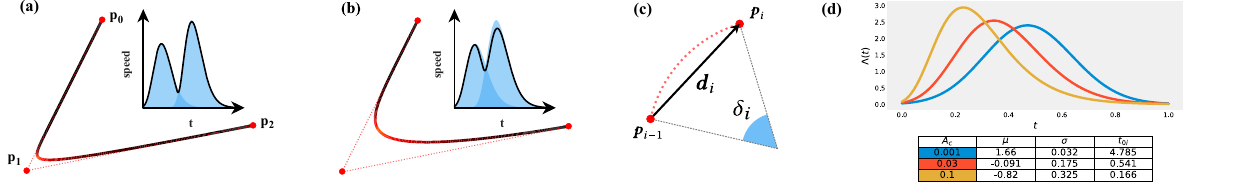} %
\caption{\textbf{(a)} and \textbf{(b)}: The effect of different time overlaps for two lognormals, where a greater time overlap results in a smoother trajectory. The motor plan is made up of virtual targets (red dots) linked by their distance separation (dotted red). In blue, the resulting speed after executing the plan and black the generated trajectory. \textbf{(c)}: Sub-movement direction $\bm{d}_{i}$ and turning angle $\delta_i$. The latter determines the internal angle of the assumed circular arc. \textbf{(d)}: Controlling lognormal shape/asymmetry using the intermediate parameter $A_{c}$ and $T=1$.  }
\label{fig:slm}
\end{figure*}

\section{Method}
Similar to existing works using the \slm model for robotics applications \cite{Berio16IROS,Wolniakowski2021}, we generate trajectories using a 2.5D extension of the model, enabling the reproduction of strokes with a varying thickness profile.

The \slm model describes the planar evolution of a handwriting trajectory through the space-time superposition of a discrete number of stroke primitives, each having the characteristic bell-shaped speed profile modeled with a two-parameter lognormal.
The spatial layout of a trajectory is described with a high-level motor plan consisting of an initial position $\bm{p}_0$ followed by a sequence of virtual targets $\left( \bm{p}_1, \dots, \bm{p}_m \right)$ (\refig{fig:slm}a). Each virtual target describes a ballistic stroke aimed along a vector \(\bm{d}_i=\bm{p}_{i} - \bm{p}_{i-1}\). Assuming movements done with rotations of the wrist or elbow, the planar component of each stroke follows a circular arc geometry determined by a turning angle parameter $\delta_{i}$ (\refig{fig:slm}c).

\subsection{Trajectory construction}
The time parametrized kinematics $\bm{x}(t), \dot{\bm{x}}(t)$ of a trajectory can be computed as linear combinations
\begin{equation}
	\begin{bmatrix}\bm{x}(t)\\ \dot{\bm{x}}(t) \\ \ddot{\bm{x}}(t)
    \end{bmatrix}
    = \begin{bmatrix}
    \bm{p}_0 \\ \bm{0} \\ \bm{0} 
    \end{bmatrix} +
	\sum_{i=1}^{m}{
    \begin{bmatrix}\bm{x}_i(t)\\ \dot{\bm{x}}_i(t) \\ \ddot{\bm{x}}_i(t)
    \end{bmatrix}
    }
  \end{equation}
  of $m$ sub-movements with position $\bm{p}_{0} + \bm{x}_{i}(t)$, velocity $\dot{\bm{x}}_i(t)$ and acceleration $\ddot{\bm{x}}_i(t)$.

  The speed profile of each sub-movement follows a lognormal function (\refig{fig:slm}d), and the overall trajectory results from the superposition of multiple sub-movements over time. Trajectory smoothness depends on the activation time and overlap of consecutive lognormals (\refig{fig:slm}b). If a new lognormal starts before the previous one is finished, the superposition creates a smooth transition between the two strokes. A greater overlap results in a smoother transition.

To maintain compatibility with an automatic differentiation pipeline, we compute the lognormal with
\begin{multline}
    \Lambda_i(t) = \frac{
		\exp\left({ - \frac{1}{2 {\sigma_i}^{2}}
(\ln(\varphi(t-t_{0i})) - \mu_i)^{2}
}\right)} {\sigma_i \sqrt{2\pi}\varphi(t-{t}_{0i})},
\label{eq:lognormal}
\end{multline}
where we use a rectifier linear unit $\varphi(t)=\text{ReLU}(t - \epsilon)+\epsilon$ with a small value $\epsilon$ to avoid values $\le 0$ in the logarithm. The parameters $t_{0i}$, $\mu_{i}$, and $\sigma_{i}$ respectively determine the activation time, delay, and response time of the lognormal.

We compute the curvilinear evolution using the integral of $\Lambda_i(u)$, which can be computed explicitly using the error function giving
\begin{equation}
  \label{eq:werf}
  w(t) = \frac{1}{2} \left[ 1 + \mathrm{erf}
\left( \frac{ \ln \left( \varphi\left( t - t_{0i} \right) \right) - \mu_i }
			{ \sigma_i \sqrt{2} } \right) \right],
\end{equation}
with $w(t) \in \left[0, 1\right]$. Each sub-movement has velocity
\begin{equation}
\dot{\bm{x}}(t) = \Lambda_i(t) \bm{H}\bm{R}[-\delta_i/2 + \delta_i w(t)]
\bm{d}_i,
\label{eq:stroke-velocity}
\end{equation}
with rotation matrix
\begin{equation}
\bm{R}[\theta] =
\begin{bmatrix}
\cos \theta & -\sin \theta & 0 \\
\sin \theta & \cos \theta & 0 \\
0 & 0 & 1
\end{bmatrix}, \quad \text{and where} %
\end{equation}
\begin{equation}
\bm{H} = \begin{bmatrix}
h & 0 & 0 \\
0 & h & 0 \\
0 & 0 & 1
\end{bmatrix} \quad \text{with} \quad
h = \text{sinc} \left( \frac{\delta_i}{2\pi} \right)^{-1}
\end{equation}
uses the normalized sine cardinal or $\text{sinc}$ function to adjust for the circular arc length, while avoiding numerical errors as the turning angle $\delta_{i}$ approaches zero.
The position along each sub-movement can be computed explicitly as a displacement with respect to the initial position $\bm{p}_{0}$ and without requiring numerical integration of \refeqn{eq:stroke-velocity} with
\begin{align}
  \bm{x}_i(t) &= \bm{Z}\bm{d}_i +
  \bm{R}[\delta_i w(t) - \delta_i]
  \left(\frac{\bm{d}_i}{2} - \bm{M}\bm{d}_i\right)
  + \bm{M} \bm{d}_i, \notag \\
  &\text{if } |\delta_i| > \epsilon,  \notag \\
  \bm{x}_{i}(t) &= w_{i}(t) \bm{d}_{i}, \notag %
              \quad \text{otherwise, with}  \notag \\
  \bm{M} &=
           \frac{1}{2\tan \frac{\delta_{i}}{2}} %
\left[
\begin{smallmatrix}
0 & -1 & 0 \\
1 &  0 & 0 \\
0 & 0 & 0
\end{smallmatrix}\right],
             \bm{Z} = \text{diag}([\frac{1}{2}, \frac{1}{2}, w_{i}(t)-0.5]).  \notag
\end{align}

The stroke acceleration is given by
\begin{multline}
  \ddot{\bm{x}}_{i}(t) = \delta_{i}{\Lambda}_{i}(t)
                         \bm{H} \dot{\bm{R}}[\delta_{i}w(t) - \frac{\delta_{i}}{2}] \bm{d}_{i} \\
                         + \dot{\Lambda}_{i}(t) \bm{H} \bm{R}[\delta_{i}w(t) - \frac{\delta_{i}}{2}] \bm{d}_{i}
\end{multline}
where
\[
  \dot{\bm{R}}[\theta] =
\begin{bmatrix}
-\sin \theta & -\cos \theta & 0 \\
\cos \theta & -\sin \theta & 0 \\
0 & 0 & 0
\end{bmatrix} \quad \text{and}
\]
\[
\dot{\Lambda}_i(t) = \Lambda_i(t) \frac{\mu_i - \sigma_i^2 - \ln (t - t_{0i})}{\sigma^2 (t - t_{0i})}.
\]

\textit{Time parametrization.} The parameters $\mu_{i}$ and $\sigma_{i}$ influence the lognormal's mode and shape, while also influencing the time occurrence of the activation parameter $t_{0i}$. To facilitate the optimization procedure that follows, we reparameterize
each submovement with: (i) a stroke duration $T_i$, (ii) a relative time offset
$\Delta t_i$ with respect to the previous stroke time occurrence and duration, and
(iii) a shape parameter $A_{c_i}$ $\in (0,1)$, which defines the skewness of
the lognormal \cite{plamondon2003kinematic}. The $\Sigma \Lambda$ parameters
$\left\{ \mu_i, \sigma_i, t_{0i} \right\}$ are then computed with
\begin{multline}
	\sigma_i = \sqrt{- \ln ( 1 - A_{c_i} ) }, \qquad \mu_i = 3 \sigma_i - \ln( \frac{ -1 + e^{6\sigma_i} }{T_i} ),\\
    t_{0i} =  t_{0i-1} + \Delta t_i \sinh (3 \sigma_i).
\end{multline}
With this intermediate formulation, the parameter $\Delta t_i$ explicitly determines the overlap between consecutive lognormals and consequently the trajectory smoothness in the proximity of a virtual target: lower values produce a greater overlap with the previous lognormal and a smoother/faster trajectory. Furthermore, a strictly positive $\Delta t_{i}$ guarantees an ordering of the lognormals, which allows us to compute the end time of the trajectory with
\begin{equation}
  T_{\text{end}} = t_{0m} + e^{\mu_{m} + 3\sigma_{m}},
  \label{eq:endt}
\end{equation}
which facilitates the definition of a minimum-time cost in the following optimization procedure.

\subsection{Differentiable rasterization}
Cubic B\'ezier curves can be defined using the Hermite formulation \cite{farinCurvesSurfacesCAGD2001} consisting of point-tangent pairs. This makes it straightforward to convert a \slm trajectory to a format compatible with DiffVG by considering the trajectory positions $\bm{x}(t)$ and velocities $\dot{\bm{x}}(t)$ and defining multiple B\'ezier curve segments between consecutive trajectory samples $t_{i}$ and $t_{i} + dt$. The control points for each segment are given by:
\begin{multline}
  \Big[
    \bm{x}(t_{i}),\; \bm{x}(t_{i}) + 3dt\dot{\bm{x}}(t_{i}),\\
    \bm{x}(t_{i}+dt) - 3dt\dot{\bm{x}}(t_{i} + dt),\; \bm{x}(t_{i}+dt) %
  \Big].
\end{multline}
We compute a fixed number $n$ of samples for each rendered trajectory using $dt=T_{\text{end}}/n$. We find that using $n=5m$ gives a sufficiently good approximation for our use cases. With this approach, the number of trajectory samples is fixed for each optimization step, while the trajectory duration changes depending on the varying values of $\Delta t_i$. 

\subsection{Executing \slm trajectories with a robot}
We first transform the trajectories from the image coordinate system (expressed in pixels) to a rectangle on the horizontal plane of a 3D coordinate system (expressed in meters). For 2.5D trajectories with a varying width profile, we convert the stroke radius to a perpendicular distance from the drawing pad with a linear mapping. We compute this mapping empirically, but this can be done automatically using visual feedback and a method such as the one proposed by G{\"u}lzow et al. \cite{gulzow2018self}.
We then resample the trajectories using a time step that ensures predefined speed and acceleration limits $v_{\max}$ and $a_{\max}$. Finally, we transform the trajectory into the drawing pad's coordinate system. We determine the drawing area and plane by manually guiding the robot to the position of four points on the drawing pad.

An optimal control problem formulation is used to plan for joint angle trajectories, by using an iterative linear quadratic regulator (iLQR) problem formulation solved with a Gauss--Newton iterative scheme \cite{Li04}, by considering end-effector orientation control and joint limits avoidance. The resulting joint angle reference trajectory is tracked by an impedance controller, resulting in a feedback controller with torque commands controlling the robot at a 1kHz rate.

\section{Results}

All the steps described to this stage, including the conversion to B\'ezier, consist of differentiable operations that are seamlessly handled by standard automatic differentiation engines such as the one provided by PyTorch \cite{PyTorch2019}. This, combined with differentiable rasterization, allows us to optimize the motor plan positions $\bm{p}_{0}, \bm{p}_{1}, \dots, \bm{p}_{m}$, the time offsets $\Delta t_{i}$, the curvature parameters $\delta_{i}$ and optionally the shape parameters $A_{c}$ with respect to a loss that trades off trajectory smoothness in a geometric sense with an objective expressed in pixel space. For all the examples that follow, we consider a compound loss

\begin{equation}
  \mathcal{L} = \lambda \mathcal{L}_{\text{smooth}} + \mathcal{L}_{\mathbb{I}},
\end{equation}
where $\mathcal{L}_{\text{smooth}}$  acts as a smoothing term by rewarding overlap between consecutive lognormals, $\mathcal{L}_{\mathbb{I}}$ is an image-based cost depending on the application and $\lambda$ is a user-defined and application-dependent scalar weight that controls the tradeoff between the two terms. We formulate the smoothing term
\[
  \mathcal{L}_{\text{smooth}} = T_{\text{end}} + w_{\sigma}\text{Var}[\bm{\Delta t}],
\]
where the second term $\text{Var}[\bm{\Delta t}]$ is the variance of the $\Delta t_{i}$ parameters and encourages uniform spacing in time between lognormals, with a weight $w_{\sigma}$ that we set to $50$ in our experiments. We observe that this is particularly important to maintain stability for long trajectories (with a high $m$), while encouraging a form of isochrony, which is consistent with empirically observed motor control principles \cite{Thomassen1985Time} and also with the Kinematic Theory, which predicts a decrease in variance as the motor control increases \cite{Plamondon2020}.
In the following sections, we demonstrate two examples of variants of $\mathcal{L}_{\mathbb{I}}$ for different applications, while always maintaining the same smoothing cost  $\mathcal{L}_{\text{smooth}}$.

We implement the optimization procedure in PyTorch and use the Adam optimizer for $300$ steps running on a NVIDIA GeForce RTX 3060. 
We test our trajectories with different painting tools, using a 7-axis Franka robot in two different configurations (inclined and horizontal drawing planes), see accompanying video.

\begin{figure}
\centering
\includegraphics[width=0.49\textwidth]{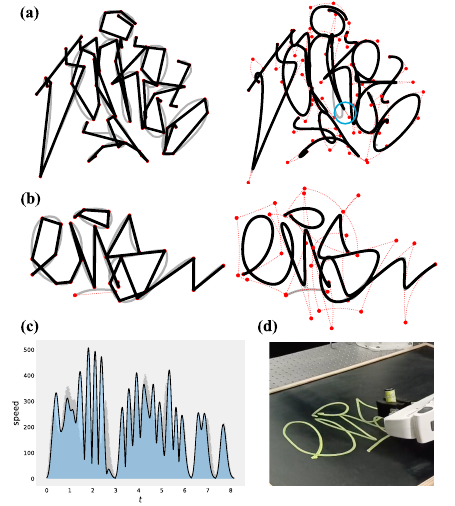} %
\caption{Image-driven trajectory reconstruction using \refeqn{eq:mse}. \textbf{(a)} Reconstructing an input trajectory (grey) starting from the motor plan (red) and trajectory (black) on the left. We start with values of $\Delta t_{i}=1$ and $\delta_i=0$ and the motor plan and trajectory initially match. Here, we optimize also the shape parameters $A_{c}$ resulting in the motor plan and trajectory on the right. In this instance, a significant reconstruction error (blue circle) occurs on the lower part of the ``K'', likely due to the presence of an inflection. \textbf{(b)} A second reconstruction keeping $A_{c}$ fixed. \textbf{(c)} the generated speed profile for \textbf{(b)} with the corresponding lognomals in blue and the original (Gaussian smoothed) speed profile in grey. The latter is not taken into consideration, but the minimum time cost produces a similar number and location of peaks. \textbf{(c)} Reproduction with a 7-axis Franka robot and a chalk marker. }
\label{fig:recon}
\end{figure}

 \subsection{Fitting trajectories to an image}
As a baseline test for our method, we test the fitting \slm trajectories to an image (\refig{fig:recon}), resulting in a procedure similar to the geometry-based fitting methods described by \cite{Berio2020Lognormality} and \cite{ferrer2018idelog}. Similar to \cite{Berio2020Lognormality}, we use tag traces from the Graffiti Analysis database \cite{GraffitiAnalysisDB}, a dataset of graffiti tags recorded with low-cost DIY tracking devices. We render the input trajectory into a black-and-white image and then create an initial motor plan for generating a \slm trajectory by applying the discrete curve evolution (DCE) polyline simplification method \cite{latecki1998discrete} to the input trace (\refig{fig:recon}). DCE simplifies polylines by identifying significant curvature extrema, with a result that is similar to the approaches of \cite{Berio2020Lognormality} and \cite{de2004saliency}. We then set
 \begin{equation}
   \mathcal{L}_{\mathbb{I}} = \sum_{s \in S} \text{MSE}(\mathbb{I}_s, \hat{\mathbb{I}}_s)
   \label{eq:mse}
 \end{equation}
to a multi-scale mean squared error (MSE) between the rendered image $\mathbb{I}$ and the target $\hat{\mathbb{I}}$. The MSE term consists of multiple downscaled and blurred versions $\mathbb{I}_s$ and $\hat{\mathbb{I}}_s$ of the input, where $s \in S$ represents different scaling factors applied to the images. 
We find that this multiscale approach improves convergence, while decreasing the risk of getting stuck in local minima. Note that the optimization infers a plausible motion from the input geometry. Different from most state-of-the-art \slm fitting methods \cite{OReilly2008,Plamondon2014}, our method does not attempt to reproduce the kinematics of the input.

\begin{figure}
\centering\includegraphics[width=0.49\textwidth]{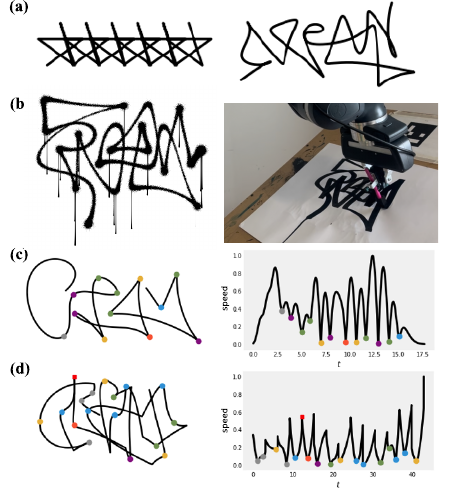} %
\caption{CLIP-guided ``pseudo''-tag generation. \textbf{(a)} Left, initial trajectory with $\Delta t_{i}=1$; right, a trajectory resulting from the optimization using \refeqn{eq:clip} with the word ``CREAM''. \textbf{(b)} Left: using the speed profile to simulate a trace made with spray paint using Gaussian samples along the trajectory; the standard deviation and density of samples are inversely proportional to speed; drips are added for effect near speed minima. Right: the same trajectory executed with a robot equipped with a sponge brush. \textbf{(c)} Speed minima (color coded) along a \slm trajectory generated with our method. \textbf{(d)} Speed minima for a piecewise cubic B\'ezier curve generated with the same method but without time-based smoothing. Note that certain curvature extrema (e.g. the one indicated with the red square) do not correspond to speed maxima, which would lead to an unnatural motion \cite{Viviani1991}.   }
\label{fig:tags}
\vspace{-0.5em}
\end{figure}

\subsection{CLIP-driven graffiti tag generation}
In a second application of our method, we test an approach similar to ClipDraw \cite{FransClipdraw2022} to guide the generation of a trace that is consistent with a given text caption or \emph{prompt} (\refig{fig:tags}) provided by the user. We set
\begin{equation}
  \mathcal{L}_{\mathbb{I}} = - {\langle f(\mathbb{I}), f(c) \rangle},
  \label{eq:clip}
\end{equation}
where $\langle \cdot \rangle$ denotes the cosine similarity measured between the embeddings $f(\mathbb{I})$ of the rendered image and the embeddings $ f(c) $ of the text caption. 
We use a caption ``$\{\text{WORD}\}$, a stylish graffiti tag'', where $\{\text{WORD}\}$ is a word we wish the trace to be similar to.
Interestingly, while the generated text is not fully legible, it possesses a structure that resembles the desired word shape and appears similar to instances of existing graffiti.

\begin{figure}
\centering\includegraphics[width=0.47\textwidth]{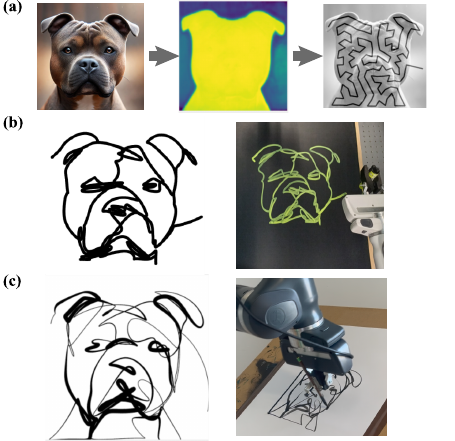} %
\caption{Single path CLIP-guided image abstraction. \textbf{(a)} Left: an input image of a dog (AI-generated); middle: saliency map for the image; right: TSP initialization path with $150$ virtual targets. \textbf{(b)} A fixed-width image abstraction and the resulting reproduction with a chalk marker. \textbf{(c)} A varying width image abstraction and the resulting reproduction using a brush dipped in Indian ink. }
\label{fig:portraits}
\end{figure}

\subsection{CLIP-driven image abstraction}
In a second CLIP-driven task, we plug our method into the \emph{CLIPasso} image abstraction method proposed by Vinker et al. \cite{vinkerCLIPassoSemanticallyAwareObject2022}, enabling the generation and robotic execution of \textit{single stroke image abstractions}. We use the L2 norm between the activations of selected internal CLIP layers $\text{CLIP}_{l}$:
\begin{equation}
    \mathcal{L}_{\mathbb{I}} = \sum_l \lambda_{l} \left\| \text{CLIP}_l(\hat{\mathbb{I}}) - \text{CLIP}_l(\mathbb{I}_{\theta}) \right\|_2^2,
    \label{eq:clip2}
\end{equation}

where $\mathbb{I}$ is the rendered image and $\hat{\mathbb{I}}$ is the target image. Similarly to CLIPasso, we start with a saliency map of the input to initialize points, but we use a different path initialization strategy (\refig{fig:portraits}a). We use a saliency map derived from the last-layer activations of the OneFormer panoptic image segmentation model \cite{oneformer} trained on the COCO dataset \cite{coco}. We then initialize a user-defined number of points using weighted Voronoi sampling \cite{secordWeightedVoronoiStippling2002} on the saliency map. Then, we connect the resulting points using a TSP \cite{kaplan2005tsp}. We find that setting all layer weights to zero with the exception of $\lambda_{2}=0.5, \lambda_{3}=0.5, \lambda_{6}=1.0$ to work well for our use cases.

\subsection{Evaluation}

\begin{figure}
\centering\includegraphics[width=0.49\textwidth]{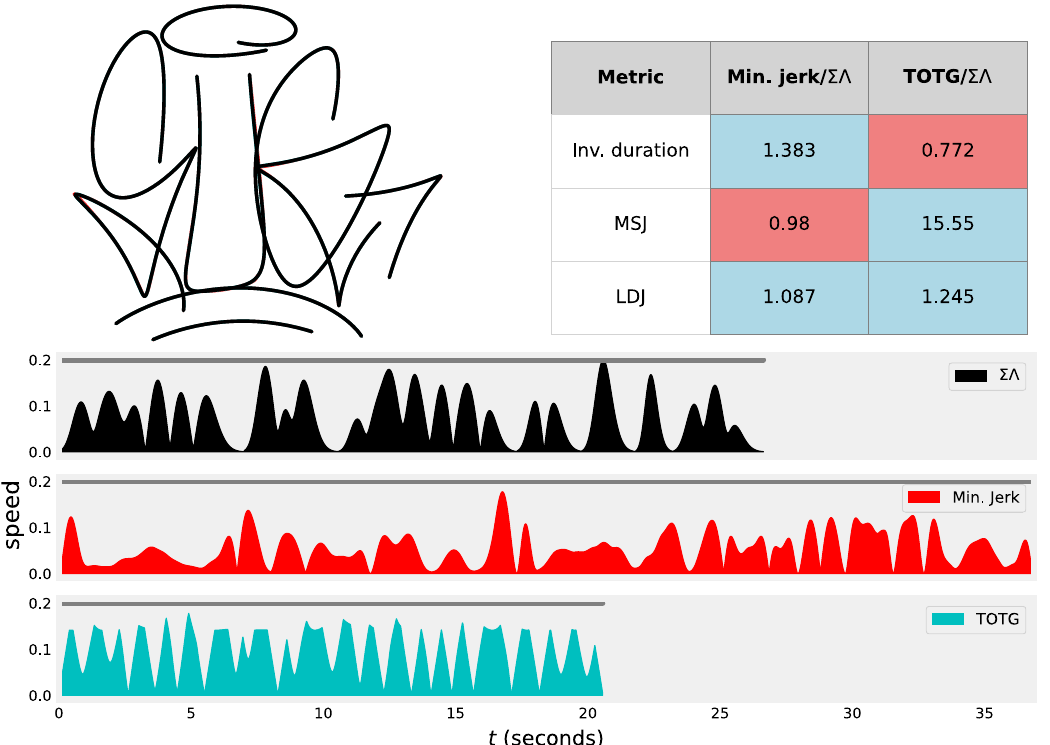} %
\caption{Comparing a reconstruction using the \slm model with minimum jerk \cite{todorovSmoothnessMaximizationPredefined1998} and TOTG \cite{Kunz2013}. Left: the \slm trajectory and its reparametrizations superposed. The trajectories are nearly identical geometrically, so only the \slm trajectory is visible in the figure. Middle: the speed profiles for \slm, minimum jerk, and TOTG parametrizations. Note that while TOTG is faster, the resulting speed has triangular and trapezoidal segments that will produce discontinuous acceleration. Right: table comparing inverse trajectory duration, mean square jerk (MSD) \cite{hoganRhythmicDiscreteMovements2007} and log-dimensionless jerk (LDJ) \cite{balasubramanianAnalysisMovementSmoothness2015} for the three parameterizations using the ratios minimum jerk/\slm and TOTG/\slm. Values $>1$ show a better performance of \slm in the metric.
}
\label{fig:comp}
\vspace{-1.2em}
\end{figure}

We performed a quantitative comparison (\refig{fig:comp}) with two different well-known trajectory parameterizations: (i) time-optimal trajectory generation (TOTG) \cite{Kunz2013}, which is widely used to plan efficient motions in robotic applications and (ii) the path-constrained minimum jerk model \cite{todorovSmoothnessMaximizationPredefined1998}, a generalization of the minimum jerk model \cite{flashCoordinationArmMovements1985} to arbitrarily complex paths, well known in the motor control literature. Both methods take a path as an input and produce a reparametrization of the path. While TOTG produces a trajectory subject to the given maximum velocity and acceleration limits, minimum jerk does not. As a result, we resample the resulting path according to the same limits.

We compared all methods in terms of trajectory duration (given the imposed limits) and smoothness according to two trajectory smoothness measures: mean squared jerk (MSJ) \cite{hoganRhythmicDiscreteMovements2007} and log-dimensionless jerk (LDJ) \cite{balasubramanianAnalysisMovementSmoothness2015}. The latter has been proposed as a robust measure of trajectory smoothness due to its invariance to scale and duration. Our results show that TOTG is faster than \slm but significantly less smooth according to both smoothness measures. At the same time, \slm trajectories are less smooth than min. jerk according to the MSJ measure but smoother according to the more recently proposed LDJ measure. This suggests that \slm provides a good tradeoff for movements that must be rapid but smooth, making it a strong candidate for applications such as drawing and writing, where these two aspects are both important.

\section{Conclusion}
We demonstrated how a 2.5D extension of the \slm model can be integrated into a DiffVG to enable image-guided optimization of smooth trajectories with kinematics that can be tracked by a drawing/painting robot. The \slm model construction provides a straightforward way to compute trajectory duration, which enables the implementation of a smoothing cost in terms of minimum time. Minimum-time is a widely used optimization objective \cite{Bobrow85} in robotic path planning, and the \slm formulation provides a representation that effectively supports this measure while producing movements that are kinematically similar to those made by a human \cite{Plamondon2014}. In quantitative comparisons, we showed that \slm provides a good tradeoff for movements that must be rapid but smooth. Evaluating the perceived naturalness and smoothness of the trajectories is an interesting avenue of future research, which links to existing work that evaluates the relation between the perceived naturalness of drawing movements and the aesthetic appreciation of the resulting traces \cite{Chamberlain2021}.

\bibliographystyle{IEEEtran} 
\bibliography{bib}

\end{document}